\documentclass{sig-alternate} 

\usepackage{times}
\usepackage{graphicx}
\usepackage{epsfig}
\usepackage{amsmath}
\usepackage{amssymb}
\usepackage{algorithm, algorithmic}
\usepackage{booktabs}
\usepackage{verbatim}

\begin{document}

% Copyright
%\setcopyright{acmcopyright}
%\setcopyright{acmlicensed}
%\setcopyright{rightsretained}
%\setcopyright{usgov}
%\setcopyright{usgovmixed}
%\setcopyright{cagov}
%\setcopyright{cagovmixed}
\CopyrightYear{2016}
%\setcopyright{acmcopyright}
\conferenceinfo{ICMR'16,}{June 06-09, 2016, New York, NY, USA}
%\isbn{978-1-4503-4359-6/16/06}\acmPrice{\$15.00}
%\doi{http://dx.doi.org/10.1145/2911996.2911997}

%

\title{GPU-FV: Realtime Fisher Vector and Its Applications in Video Monitoring}
%\subtitle{[Extended Abstract]
%\titlenote{A full version of this paper is available as
%\textit{Author's Guide to Preparing ACM SIG Proceedings Using
%\LaTeX$2_\epsilon$\ and BibTeX} at

\numberofauthors{5} %  in this sample file, there are a *total*
\author{
% 1st. author
\alignauthor
Wenjing Ma \\
\affaddr{State Key Laboratory of Computer Science,}\\
       \affaddr{Laboratory of Parallel Software and Computational science,}\\
       \affaddr{Institute of Software, Chinese Academy of Sciences}\\
       \email{wenjing@iscas.ac.cn}
% 2nd. author
\alignauthor
Liangliang Cao\\
\affaddr{Yahoo Labs}\\
\affaddr{New York City}\\
\affaddr{New York, USA}\\
       \email{\mbox{liangliang@yahoo-inc.com}}
% 3rd. author
\alignauthor Lei Yu\\
      \affaddr{Laboratory of Parallel Software and Computational science,}\\
       \affaddr{Institute of Software, Chinese Academy of Sciences}\\
       \email{yulei@iscas.ac.cn}
\and  % use '\and' if you need 'another row' of author names
% 4th. author
\alignauthor Guoping Long\\
      \affaddr{Laboratory of Parallel Software and Computational science,}\\
       \affaddr{Institute of Software, Chinese Academy of Sciences}\\
       \email{guoping@iscas.ac.cn}
       % 5th. author
       \alignauthor Yucheng Li\\
       \affaddr{Laboratory of Parallel Software and Computational science,}\\
       \affaddr{Institute of Software, Chinese Academy of Sciences}\\
       \email{yucheng@iscas.ac.cn}
}
% There's nothing stopping you putting the seventh, eighth, etc.
% author on the opening page (as the 'third row') but we ask,
% for aesthetic reasons that you place these 'additional authors'
% in the \additional authors block, viz.
%\additionalauthors{Additional authors: John Smith (The Th{\o}rv{\"a}ld Group,
%email: {\texttt{jsmith@affiliation.org}}) and Julius P.~Kumquat
%(The Kumquat Consortium, email: {\texttt{jpkumquat@consortium.net}}).}
% Just remember to make sure that the TOTAL number of authors
% is the number that will appear on the first page PLUS the
% number that will appear in the \additionalauthors section.

\maketitle
\begin{abstract}
    Fisher vector has been widely used in many multimedia retrieval and visual recognition applications with good performance.
  However, the computation complexity prevents its usage in  real-time video monitoring.
  In this work, we proposed and implemented GPU-FV, a fast Fisher vector extraction method with the help of modern GPUs.
  The challenge of implementing Fisher vector on GPUs lies in
  the data dependency in feature extraction and expensive memory access in Fisher vector computing.
  To handle these challenges, 
  we carefully designed GPU-FV in a way that utilizes the computing power of GPU as much as possible, and applied optimizations such as loop tiling to boost the performance. 
  GPU-FV is about 12 times faster than the CPU version, and 50\% faster than a non-optimized GPU implementation. 
  For standard video input (320*240), GPU-FV can process each frame  within 34ms on a model GPU. Our experiments show that
  GPU-FV obtains a similar recognition accuracy as traditional FV on VOC 2007 and Caltech 256 image sets. We also applied  GPU-FV for realtime video monitoring tasks and found that GPU-FV outperforms a number of previous works. Especially, when the number of training examples are small, GPU-FV  outperforms the recent popular deep CNN features borrowed from ImageNet. 
\end{abstract}

\keywords{Fisher vector; image classification; real-time; event detection; GPGPU}

%ACM proceedings; \LaTeX; text tagging}

\section{Introduction}

Many recent research show that Fisher vector \cite{FV-ijcv13} is a very useful representation of images, which obtains state-of-the-art results  in a number of applications,
including image retrieval \cite{perronnin2010large, douze:inria-00566293}, art classification~\cite{MensinkICMIR2014}, relevance feedback of videos \cite{Feedback}, event recounting in videos~\cite{Sun:2014ICMR}, texture recognition \cite{Cimpoi14}, face verification \cite{fv-face-bmvc13}, object detection  \cite{fv-detection-cvpr13, flair-cvpr14}, and
fine grained image recognition\cite{gavves2013fine,fineClass}.
Although the recent deep convolutional neural network \cite{DBLP:conf/nips/KrizhevskySH12} outperforms Fisher vector in large scale visual recognition challenges such as ImageNet LSVRC \cite{imagenet_cvpr09}, the training process of deep CNN usually requires a lot of training examples. In scenarios where the training set is in small or middle scale, Fisher vector is still a very attractive choice, especially when the problem lies in a different domain from the existing large scale dataset. 

However, Fisher vectors pose an even heavier computational
demand,  which limits its usage in real world applications. It takes about 2-5 seconds for a typical implementation of Fisher vector to classify a normal sized image. Such a slowness prevents Fisher vectors from many real applications.
To bridge the gap, 
this paper proposes GPU-FV, an efficient implementation of Fisher vector with the help of GPUs~\footnote{The code can be downloaded from the following link https://bitbucket.org/mawenjing/gpu-fv}.

%we shall talk about GPU here....

%In this paper, we proposed an efficient implementation of Fisher vector with the help of GPUs. It is not straightforward to implement Fisher vector in GPUs due to the ***. To overcome this challenge, we not only design efficient kernels to explore  the parallel nature in SIFT extraction, but also formulate GMM estimation in a form that can utilize the limited bandwidth and memory in modern GPUs. We propose several techniques to optimize the feature extraction speed: (1) Loop tiling, (2) early thresholding, (3) thresholding and (4) using 8 scales for classification. We demonstrate that with our optimization technique, the Fisher vector can speed up a naive GPU implementation by ***\%. 

It is not trivial to write efficient GPU code for Fisher vector, because of the data dependency in dense SIFT, and the complicated memory access patterns in GMM estimation.
To overcome this challenge, we not only design efficient kernels to explore  the parallel nature in SIFT extraction, but also formulate GMM estimation in a way that can utilize the limited bandwidth and memory in modern GPUs. We propose several techniques to optimize the feature extraction speed: (1) Loop tiling, (2) early termination, and (3) vectorization, and (4) using 8 scales for dense SIFT exatraction. We demonstrate that  our optimization technique can speed up a naive Fisher vector implementation on GPUs by more than 20 times. 

Some techniques in optimizing the GPU computation bring approximation of the computation. To evaluate the effects of these approximations, we compare our GPU implementation with the original Fisher vector on two standard datasets: PASCAL VOC 2007 and Caltech 256. The results show that our approximate obtain a similar accuracy but with a much higher speed. We believe such evaluation provides strong evidence that our GPU based Fisher vector can be used for many applications. 

Our GPU-FV system can process a 320 * 240 image in 34ms. Such an efficiency makes it possible to employ Fisher vector to realtime applications. To demonstrate it, this paper shows an example of applying an abnormal event detector in realtime surveillance videos, and an example of detecting baby's laugh in a video. 
In surveillance and video recognition, existing solutions are very slow and it is difficult to apply them to online processing scenarios. However, this paper demonstrates that our new GPU based Fisher vector can model complicated subjects and obtain an accuracy comparable with the the state-of-the-art.% Another example in the paper is detecting baby's laugh from a video. %TODO: finish it.

%the efficient approaches are either based on offline processing or limited to  efficient but simple approaches for video analysis. 

%For example, consider the problem of recognizing the baby's smile in the video. We show our GPU-based FV can be used for (1) efficiently applying a pre-trained smile detector in the video sequence, and (2) training a smile detector online by providing a few positive examples. 
%The efficiency and accuracy can be compared with the state-of-the-art visual recognition method. 

\section{Related Works}
%\textbf{GPU-based related works}
GPUs have been evolving fast, and have been applied in high performance computing successfully.
%with transistor
%counts doubling every few months.
However, it remains unclear what are the  bottlenecks in accurate visual categorization with the Fisher vector model on GPU.

SURF is an optimized robust feature extraction system~\cite{SURF2008}.
Cornelis and Van
Gool \cite{gpu-for-surf08} implemented SURF on the GPU (Graphics
Processing Unit) and obtained an order of magnitude
speedup compared to a CPU implementation.
Extracting SIFT descriptors on GPU has been studied by other researchers~\cite{MVA11Sinha,HPCC13Wang}. Recent efforts were also made on accelerating Dense SIFT computation~\cite{GeoComputation2015}.
SVM model training, have been independently studied on the GPU before \cite{gpu-svm-icml08}.
To address the bottlenecks in accurate visual categorization systems, Sande et.al \cite{vandeSandeITM2011} did a detailed analysis, and proposed
%proposed (1) an analysis of the bottlenecks in
%accurate visual categorization systems and, to address these
%bottlenecks, (2)
two GPU-accelerated algorithms, GPU vector
quantization and GPU kernel value precomputation, which
results in a substantial acceleration of the complete visual
categorization pipeline. However, their method does not involve Fisher vector encoding.

Efforts have been made to reduce the storage and computation overhead of Fisher vector~\cite{perronnin2010large}, by compressing the Fisher vector, with some loss of precision.
The widely used Vlfeat package ~\cite{vedaldi08vlfeat} provides a wonderful implementation of the Fisher vector, along with other popular computer vision algorithms. However, there was no GPU-based implementation in this package.
A few years back, some researchers implemented Fisher vector on GPU~\cite{petras2011gmm}, but it is on a modified algorithm with hierarchical GMM model, and the accuracy is lower than the state-of-the-art. 

The problem of abnormal event recognition in videos has attracted many attentions
\cite{zhao2011online} \cite{kratz2009anomaly} \cite{lu2013abnormal}. However, 
%to the best of our knowledge, there has been no previous work of applying Fisher vector to this problem due to its slowness.
the approaches listed above did not use Fisher Vector due to its slowness.
Chen et al. \cite{Chen:2013:SFV} used MoSIFT and Fisher Vector for event detection, and obtained good performance on TRECVID data. However, their work did not consider how to speed up local feature extraction or Fisher vector encoding. As a result,  their method relies on significant subsampling of one from 30/60/120 frames and the time of encoding such a frame is about 0.4 second (with feature extraction it will be longer). We believe our work in this paper can be easily employed by the framework \cite{Chen:2013:SFV} and provide similar speed up.
%Chen et al. used MoSIFT and Fisher Vector for event detection, and were able to extract Fisher Vector 10 times faster than real-time~\cite{Chen:2013:SFV}.
%However, to obtain fast speed, they have to subsample the video once per 120 frames (they selected the best result from tests on 30, 60, and 120 frame sliding windows), implying that the time to encode one frame is about 0.4s. Besides, the time they reported excluded the time for generating the local feature, which is also a time-consuming component in processing each frame.
Oneata et al. also used Fisher Vector in action localization and event recognition~\cite{oneata2013ICCV}, at a speed 2.4 times slower than real-time. 
In this paper, we demonstrate that with GPU based Fisher Vector, we can handle some abnormal event recognition very well at a realtime speed.
%, which is even faster than many of previous works. 

In recent years, 
deep neural network have enjoyed a remarkable success as efficient and effective in a number of visual recognition tasks~\cite{NIPS2014,DuTran15ICCV}. Especially, Razavian et al.
\cite{DBLP:journals/corr/RazavianASC14} showed that by simply borrowing  the CNN-based AlexNet model \cite{DBLP:conf/nips/KrizhevskySH12} trained for ImageNet, a SVM model using CNN features can obtain the state-of-the-art in many applications. The deep CNN features could be sped up significantly by GPUs. We believe Fisher vector can be sped up with the same hardware, and in this paper we show that GPU-FV can outperform deep CNN features in some applications with limited amount  of training samples.

\section{GPU-FV}
  %Our Efficient Implementation}

In this section, we will first introduce the Fisher vector algorithm and then explain our GPU-based implementation in detail.

\subsection{Overview of Fisher Vector}
\label{sec:basic}
There are two main computation components when generating Fisher vector for a picture.
The first component is extracting the dense SIFT descriptors. 
An image is scaled to different sizes.
Descriptors are extracted for all the scales.
%, and concatenated into a feature vector.
For each scale, the gradients of the pixels fall into  8 orientation bins.
Then, the convolution kernels are applied.
After that, each value in the 8 orientation bins is multiplied with the weights computed for each of the 16 spatial bin (4 on X and 4 on Y dimension).
%and sift descriptors are extracted on each scale. 
Therefore, a 128 dimension descriptor(8*4*4) is generated for each pixel.
Then, by lowering the dimension to $m$ ($m<$128) with PCA and adding the normalized X and Y axis, a descriptor with dimension $M$=$m$+2 is generated for a pixel.
The descriptors are extracted with a stride of $S$ on X and Y dimension, which means we get a descriptor in every $S^{2}$ pixels.

The second component is using the descriptors to generate the Fisher vectors with the GMM components trained beforehand.
With the $means$, $covariances$, and $priors$ values of GMM, the dense SIFT descriptors are used to generate a Fisher vector for an image.
The algorithm is described in Algorithm~\ref{alg:fv}~\cite{FV-ijcv13}.
Generating the Fisher vector includes two phases.
In the first phase, we get the posteriors for each descriptor regarding the GMM components.
In the second phse, the Fisher vector for the image is generated, represented as the concatenation of vectors $V$ and $U$ in Algorithm~\ref{alg:fv}.
$V$ and $U$ are of size $M$*$N$, where $M$ is the dimension of a descriptor, and $N$ is the number of GMM components.
%Since the dimension of the dense SIFT descriptor and the number of components are fixed, the Fisher vectors of different images would have the same length.
\begin{algorithm}[htb]
\caption {Fisher Vector Encoding Algorithm} 
\label{alg:fv}
\algsetup{indent=1em}
\begin{algorithmic}[1]
\renewcommand{\algorithmicrequire}{\textbf{Input:}}
\renewcommand{\algorithmicensure}{\textbf{Output:}}
\renewcommand{\algorithmicforall}{\textbf{foreach}}

\REQUIRE{Dense SIFT descriptors $data$}
\REQUIRE{GMM model: $means$, $covariances$, $priors$}
\ENSURE{Fisher Vector: {$V$,$U$}}
\STATE {Compute $\sqrt{{\sigma}^{-1}}$}
\FOR {$i=0$ to $num\_descriptors$}
\FOR{$j=0$ to $num\_components$}
\STATE {$t$=distance(${data[i]}, {means_j}, {covariances_j}$);}
\STATE {Compute $posteriors_{i,j}$ with $temp$;}
\STATE{$maxPost$ = max\{$maxPost$, $posteriors_{i,j}$\};}
\ENDFOR
\FOR{$j=0$ to $num\_components$}
\STATE{$posterior_{i,j}$ = $e^{posterior_{i,j}-maxPost}$;}
\STATE{$sum$ += $posterior_{i,j}$;}
\ENDFOR
\FOR{$j=0$ to $num\_components$}
\STATE{$posterior_{i,j}$ = $posterior_{i,j}/sum$;}
\ENDFOR
\ENDFOR
\FOR{$i=0$ to $num\_descriptors$}
\FOR{$j=0$ to $num\_components$}
\IF{$posterior_{i,j}>$threshold}
\FOR{$k=0$ to $DIMENSION$}
\STATE{$U_{j,k}$+=$(data[i][k]-means_{j,k})*\sqrt{{\sigma}^{-1}}*posteriors_{i,j}$;}
\STATE{$V_{j,k}$+=$(((data[i][k]-means_{j,k})*\sqrt{{\sigma}^{-1}})^2-1)*posteriors_{i,j}$;}
\ENDFOR
\ENDIF
\ENDFOR
\ENDFOR
%\STATE {Compute $U$ and $V$ with $posteriors$}
%TODO: finish the algorithm description
%\STATE Compute
\end{algorithmic}
\end{algorithm}

\subsection{The Parallelization Scheme on GPU}

In this section, we describe the implementation details of the GPU-FV system.

\textbf{Terminology} First, we introduce some terminology to be used when describing GPU programs.
A GPU kernel is a function launched by the CPU and run on the GPU.
A kernel is run by a $thread\ grid$, which consists of a bunch of $thread\ blocks$. Each $block$ is comprised with a certain number of threads. 
Threads in the same block run on the same multiprocessor in a GPU, and share a piece of scratchpad memory, called $shared\ memory$, which could be read and written by the program, and is much faster than the GPU device memory.
Threads are scheduled as $warps$, which is a bunch of threads that run in a SIMD fashion.
With conditional branches such as ``if...else...'', if the condition varies among threads in the same warp, the whole warp must go through both branches.

%Vlfeat ~\cite{vedaldi08vlfeat}

\begin{figure}[ht]
\includegraphics[width=11cm]{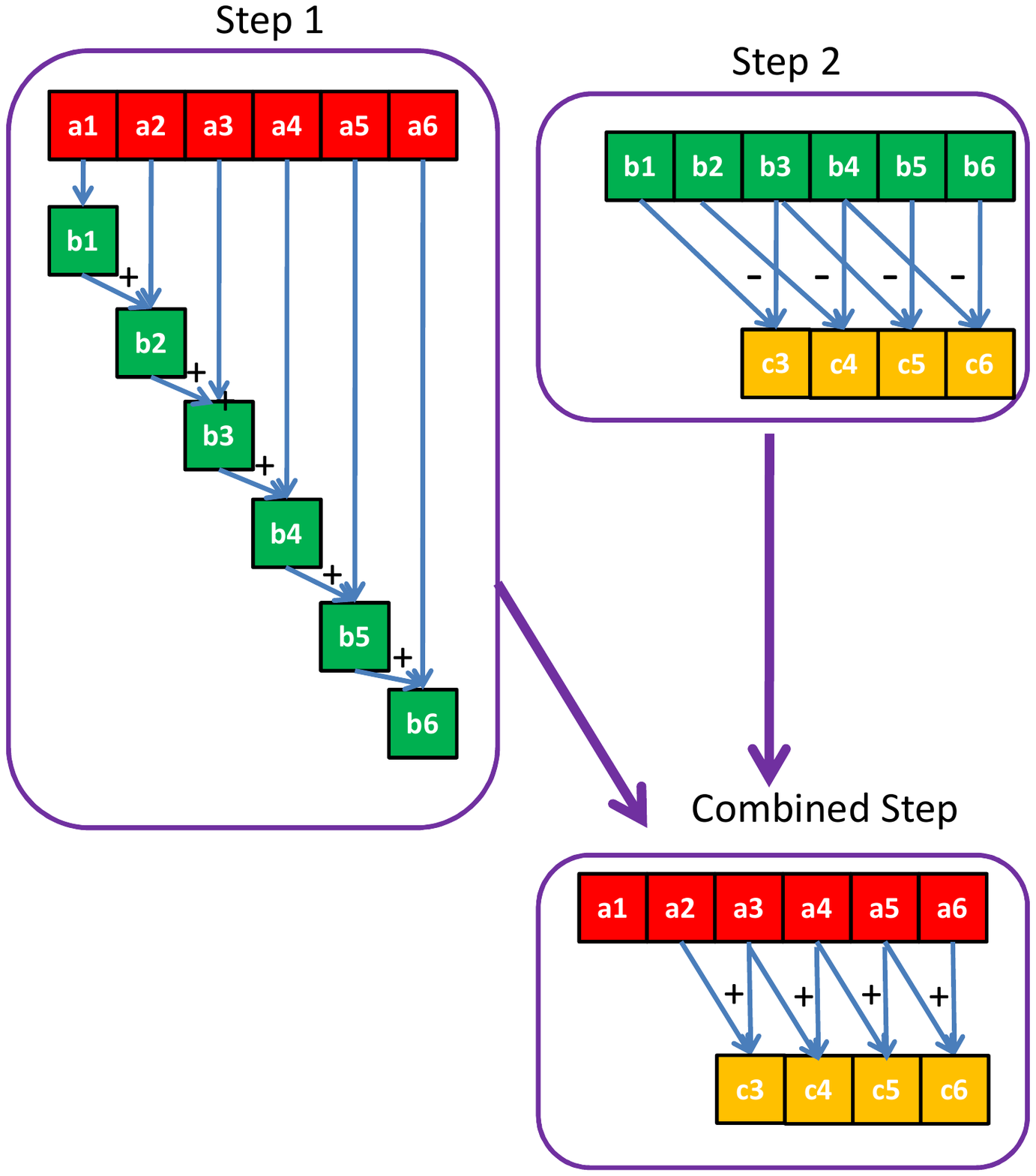}
\caption{Combining two steps in Dense SIFT}
\label{fig:steps}
\end{figure}

\textbf{Dense SIFT on GPU} 
There are two types of dense SIFT extracting methods: flat window and Gaussian window. In flat window dense SIFT, the triangular convolution kernels are used, and the convolution routine need to be called only twice (once on each dimension) for the 8 gradient bin, therefore reducing the amount of computation.
%The image classification application in Vlfeat uses the flat window scheme to extract dense SIFT descriptors.
To leverage this optimization, we implemented our GPU code based on the flat window dense SIFT algorithm.
However, the original CPU implementation cannot be changed to GPU code directly because of its sequential nature.
The triangular convolution involves 4 steps: (1) integrate backward the column, (2) compute the filter forward, (3) integrate forward the column, and (4) compute the filter backward.
The first and third steps are accumulations, while the second and the fourth steps are subtraction operations.
Each step is a $scan$ operation in nature, as shown in Step 1 in Figure~\ref{fig:steps}.
Because the computation of $b[i]$ depends on the result of $[i-1]$ computed in the last iteration, the computation of $b[i]$ and $b[i-1]$ cannot be done in parallel.
$Scan$ operation need aggressive optimization when implemented on GPU to gain performance benefits~\cite{DBLP:conf/ppopp/YanLZ13}, and $scan$ with subtraction is even harder to be implemented on GPU. 

With careful examination, we found that there is useless computation in this process, because some data are added (in step 1 or 3), then subtracted (in step 2 or 4).
%Therefore, we can combine the 4 steps, by combining the first two steps, and the last two steps.
Therefore, we revised the algorithm into a 2-step procedure, by combining the first two steps, and the last two steps.
In each of the two steps, GPU threads sum up only the needed \textit{f} data for each output point, where \textit{f} is the filter size.
Figure~\ref{fig:steps} shows this combination. The red squares are inputs to the first step.
The green squares are the results after the adding operation in Step 1.
The yellow squares are the results after two steps, which is the $scan$ operation with subtraction.
The window size is 2.
By combining Step 1 and Step 2, we got a simplified computation pattern, as shown in the ``Combined Step''.
Now the subtraction operation is not required any more.
This algorithm could run efficiently on GPU.
We spread the threads in each thread block on the columns of an image, and let each thread block process one column of the image.
%Thus, each thread does the convolution for a pixel by doing two rounds of adding operations.
%In each round, $K$ adding operations is done, where $K$ is the size of the convolution kernel.
Thus, each thread does the convolution for a pixel by doing two rounds of adding operations, with $f$ adding operations in a round.
Threads in the same block use the shared memory to buffer the data in the same column of the image, to accelerate data access.

We also optimized the GPU computation by removing computation for unused pixels.
As mentioned in Section~\ref{sec:basic}, the descriptors are extracted with a step of 4 on X and Y dimension, which means we need only 1 descriptor in every 16 pixels.
In the original code, all the pixels are processed with both of the two convolution operations.
In the GPU code, we reduced computation time by computing only the required data in the second convolution.

Generating the dense SIFT descriptors need preprocessing of images, such as image resizing, computation of gradients, normalization, etc.
These steps are easy to be implemented as GPU kernels.
With the preprocessing done on GPU, we were able to avoid extra data copy from host memory to GPU memory.
%Though some threads are idle when processing the points close to the boundary, the idle time is insignificant, since $f$ is always much smaller than the size of each dimension of an image.
%Since the filter size is small, we reduce the idle time of threads significantly. 

\begin{algorithm}[htb]
\caption {GPU code for Phase 1} 
\label{alg:gpu1}
\algsetup{indent=1em}
\begin{algorithmic}[1]
\renewcommand{\algorithmicrequire}{\textbf{Input:}}
\renewcommand{\algorithmicensure}{\textbf{Output:}}
\renewcommand{\algorithmicforall}{\textbf{foreach}}
\REQUIRE{Dense SIFT descriptors $data$}
\REQUIRE{GMM model: $means$, $covariances$, $priors$}
%\ENSURE{Fisher Vector: {$V$,$U$}}
%\ENSURE{posteriors}
%\STATE {Launch a GPU kernel to compute $\sqrt{{\sigma}^{-1}}$;}
%\STATE {//StarKernel for Phase 1}
\STATE {//Each block processes one descriptor at a time}
\FOR {$i=0$ to $num\_descriptors$ }
\STATE{Load all dimensions of $data[i]$ to shared memory;}
\STATE {Synchronize();}
\FOR{$j=0$ to $num\_components$}
\STATE{//Thread $k$ processes $data[i][k]$}
\STATE {$t$=distance($data[i][k]$, $means_{j,k}$, $covariances_{j,k}$ ); }
\STATE {Sum up $t$ from all the threads in the block in a tree structure;}
\STATE {Synchronize();}
\IF{ThreadID==0}
\STATE {Compute $posteriors_{i,j}$ with $t$;}
\STATE{$maxPost$ = max\{$maxPost$, $posteriors_{i,j}$\};}
\ENDIF
\STATE {Synchronize();}
\ENDFOR
\STATE {//Each thread processes one component at a time}
\FOR{$j=0$ to $num\_components$ }
\STATE{$posterior_{i,j}$ = $e^{posterior_{i,j}-maxPost}$;}
\STATE{$sum$ += $posterior_{i,j}$;}
\ENDFOR
\STATE {Synchronize();}
\STATE {Sum up  $sum$ with all the threads in the block in a tree structure;}
\STATE {Synchronize();}
\STATE {//Each thread processes one component at a time}
\FOR{$j=0$ to $num\_components$ }
\STATE{$posterior_{i,j}$ = $posterior_{i,j}/sum$;}
\ENDFOR
\ENDFOR
\end{algorithmic}
\end{algorithm}

\begin{algorithm}[htb]
\caption {GPU code for Phase 2} 
\label{alg:gpu2}
\algsetup{indent=1em}
\begin{algorithmic}[1]
\renewcommand{\algorithmicrequire}{\textbf{Input:}}
\renewcommand{\algorithmicensure}{\textbf{Output:}}
\renewcommand{\algorithmicforall}{\textbf{foreach}}
\STATE {//Each block processes one descriptor at a time}
\FOR{$i=0$ to $num\_descriptors$ }
\STATE {Load all dimensions of $data[i]$ and $posterior_i$ to shared memory;}
\FOR{$j=0$ to $num\_components$}
\STATE {//Thread $k$ processes $data[i][k]$}
\IF{$posterior_{i,j}>threshold$}
\STATE{$U_{j,k}$+=$(data[i][k]-means_{j,k})*\sqrt{{\sigma}^{-1}}*posterior_{i,j}$;}
\STATE{$V_{j,k}$+=$(((data[i][k]-means_{j,k})*\sqrt{{\sigma}^{-1}})^2-1)*posterior_{i,j}$;}
\ENDIF
\ENDFOR
\ENDFOR
\STATE {//Start a new kernel for accumulation of $U$ and $V$}
%\STATE {Add values in $U$ and $V$ computed by each thread block}
\STATE {//Each block processes $U$ and $V$ for one GMM component}
\FOR{$i=1$ to $num\_blocks$ }
\STATE {//Thread $k$ processes $U_{j,k}$ and $V_{j,k}$}
\STATE {$U_{j,k}$+=the $i$th copy of $U_{j,k}$}
\STATE {$V_{j,k}$+=the $i$th copy of $V_{j,k}$}
\ENDFOR
%\STATE {Compute $U$ and $V$ with $posteriors$}
%TODO: finish the algorithm description
%\STATE Compute
\end{algorithmic}
\end{algorithm}

\textbf{Generating Fisher Vector on GPU} 

The GPU implementation of the two phases  is shown in Algorithm~\ref{alg:gpu1} and \ref{alg:gpu2} respectively.
In Phase 1, the posterior values are calculated for each descriptor.  Phase 2 generates Fisher vector based on the posteriors. Therefore, both phases involve an outer loop on the number of descriptors. 
So it is natural to parallelize the outer loop (the loop on the number of descriptors) at block level on GPU.
It means thread block $i$ processes $data[i*CHUNK]$ to $data[(i+1)*CHUNK]$ by looping through the $CHUNK$ descriptors.
An important factor affecting performance is the choice of $CHUNK$. We will explain how we choose this value later in this section.
Now let us look at the implementation of Phase 1 first.
Algorithm~\ref{alg:gpu1} describes the implementation of the main kernel on GPU, and before this function starts, $\sqrt{{\sigma}^{-1}}$ has been computed with a simple GPU kernel.
In this phase, the most time consuming computation is calculating the distances between the descriptor and each GMM component. 
%Therefore, the distance value is computed by all the threads in a block, which means each thread processes one element in the descriptor.
%The computation of the distance is parallelized, and done by all the threads in a block in collaboration.
We parallelized this computation at thread level.
For each GMM component, every thread processes one dimension of the descriptor, and obtain the distance value in that dimension.
%Since each descriptor has a dimension of 82, we use 96 threads for each block, which is the closest multiple of 32 (the number of threads in a warp on GPU). 
The results of all the threads are summed up by a tree structured reduction operation, to obtain the final distance value of this descriptor.% by summing up distance values of all the dimensions.
(Intuitively, we could let each thread process all the dimensions for each GMM component,  therefore avoiding the need for reduction on the computation of distances. However, we found that the parallelization on the inner loops is more beneficial, because it enables coalesced access to global memory, and avoids bank conflicts when accessing shared memory.)

At the end of the reduction process, Thread 0 would have the distance value for the descriptor.
Then Thread 0 computes the posterior value for the descriptor, and updates $maxPost$.

%In our framework, we used the same settings as the image classification application in Vlfeat, which is 256 GMM components, and 82 dimensions dense SIFT descriptor.
%Intuitively, we could parallelize the first loop in Algorithm~\ref{} at component level, which means each thread processes one GMM components.
%It favors the setting of 256 threads in a block on GPU, since no thread is idle in the process, except when getting the maximum value.
%However, we found that parallelizing the inner loop, which is the loop on dimension, gains more performance benefit, though some threads are idle.
%This is because

%Another loop is needed to sum up the posterior values for all the components.
%TODO: need to explain it more clearly.
After that, the posteriors are updated with the exponential operation, as in Line 17 in Algorithm~\ref{alg:gpu1}.
Then, to get the summation of the posteriors for all the components, a new round of reduction is conducted.
Finally, we are able to get the normalized posterior values by dividing the posteriors with the sum.

%\begin{figure}[ht]
%\includegraphics[width=7cm]{reduction.pdf}
%\caption{Reduction operation on GPU}
%\label{fig:reduction}
%\end{figure}

In the second phase, we compute the Fisher vector (\textit{U} and \textit{V}) with the posteriors. 
Similar to the first phase, each thread block processes $CHUNK$ descriptors assigned to it in a loop.
Every thread computes one dimension in \textit{U[j]} and \textit{V[j]} at a time.
As shown in Line 7 and 8 in Algorithm~\ref{alg:gpu2}, the calculated values are used to update \textit{U} and \textit{V}.
A challenge in updating \textit{U} and \textit{V} values on GPU is that all the blocks are updating the same elements in \textit{U} and \textit{V}.
To solve this problem, we keep one copy of \textit{U} and \textit{V} for each block, and every blocks only updates its own copy. 
%having fixed length regardless of the number of descriptors, 
%In this phase, since the $U$ and $V$ values need to be summed from each descriptor, we let each block processes more descriptors.
%Therefore, we used more blocks than in phase 1.
%Since each block  processes D*TILE data, the thread grid configuration is different from the first phase. Therefore, we launch a new kernel for it. 
When this kernel finishes, each block has computed the values of \textit{U} and \textit{V} regarding the posteriors that it process.
Then, we launch a new kernel to sum up the $U$ and $V$ copies updated by each block, to obtain the final Fisher vector.%256*82*2 dimension Fisher vector.

The descriptor and the posterior values are buffered in shared memory, since they can be reused when processing each component, as Line 3 in Algorithm~\ref{alg:gpu2}.
%\textit{U[i]} and \textit{V[i]} are the final descriptors, with values accumulated from all the descriptors.

Now let us revisit the problem of choosing the values of $CHUNK$. 
We used different $CHUNK$ values for the two phases.
This is because Phase 2 needs to accumulate the $U$ and $V$ values calculated from each descriptor. Having a large $CHUNK$ implies fewer thread blocks, because each block processes more data.
Therefore, we will have less copies of $U$ and $V$, reducing the overhead of summing up values from different blocks.
On the other hand, using small $CHUNK$ number for Phase 1 will enable better load balance and higher level of parallelism, so we chose a smaller $CHUNK$ size for Phase 1.
Different $CHUNK$ values imply different thread block numbers, therefore we launch a separate kernel for each phase.
By testing with different values, we chose $CHUNK$=4 in Phase 1, and $CHUNK$=192 in Phase 2 in our experiments.
%TODO: need to specify the CHUNK numbers.

%\textbf{Compare CPU implementation and GPU implementation}
\subsection{Optimizing   GPU Implementation}
\label{sec:opt}
\textbf{Loop tiling} As a common optimization in CPU code, loop tiling turned out to be an effective optimization for GPU code also.
Normally, loop tiling improves performance by reducing the number of  conditional checking.
In our case, this technique provides more benefit by avoiding a number of synchronizations among threads in a block.
As shown in Algorithm~\ref{alg:gpu1}, we need synchronizations before and after thread 0 computes the posteriors. Besides, there is a synchronization operation after reduction on every level.
%in Figure~\ref{fig:reduction}.
Suppose there are 10 synchronizations in the loop body.
If we tile the loop by 4,  the total number of synchronizations in 4 iterations is still 10, therefore the number of synchronizations is reduced by 75\%.
%, as shown in Algorithm~\ref{}.

In the first phase, we tiled the outer loop by 4, which means each block processes 4 descriptors at a time.
The inner loop, which is on the number of components, is tiled by 2.
The loops in the second phase are also tiled. The outer loop (on the number of descriptors) is tiled by 2, and the inner loop (on the number of components) is tiled by 4.
%Since the tiling schemes are different in the two phases, and the two phases require different data buffering schemes, we launch a separate kernel for each phase.

\textbf{Early termination of an iteration} In the second phase, we were able to apply early termination of an iteration by checking conditions in a clustered way.
In the original algorithm, if the posterior of a certain point is less than a threshold, the value does not need to be added, as Line 18 in Algorithm~\ref{alg:fv}.
In the tiled GPU code, each block processes 8 posteriors in one iteration,
so we load the $means$ value to shared memory for computation before processing the descriptors, since the $means$ values could be reused.
%it implies that each thread need to check 8 times in each iteration.
However, we noticed that the posterior values are very sparse, therefore, if all the 8 values in an iteration are below the threshold, we can simply skip the iteration, as shown in Algorithm~\ref{alg:optimized}.
In this case, we avoided the loading of $means$ for this iteration, and we combined the 8 branches into one branch, which is beneficial to GPU code because of its SIMD nature.
%This is beneficial especially when all the threads in one warp could skip the whole iteration.
%In GPU, threads in one warp execute in a SIMD fashion, so the whole warp can skip an iteration only when all the threads get the same conditional values.

\textbf{Vectorization} Vectorization is an importation optimization for modern GPU.
With loop unrolling, each thread is processing multiple data elements, therefore we pack the operation on multiple data elements into a vector operation.
For example, when computing the distance between a data point and a component, the loop is unrolled with a factor of 4. Therefore we use vector operations on vectors with 4 float point data.
By applying vectorization to the computation of posteriors on GPU, we were able to further improve the performance. 

\begin{figure}[ht]
\includegraphics[width=7cm]{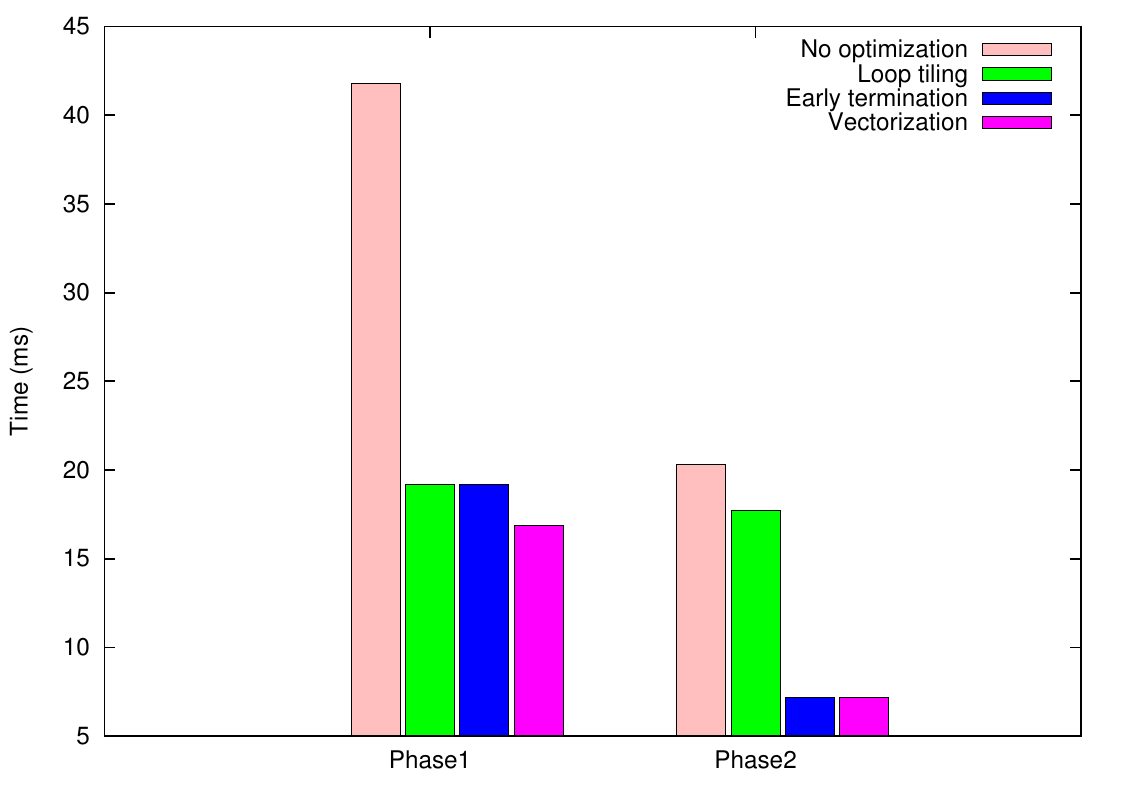}
\caption{Performance improvement with the optimization}
\label{fig:optimization}
\end{figure}

Figure~\ref{fig:optimization} shows the effectiveness of the optimization.
The bars with ``no optimization'' means the code without optimization.
The bars labeled ``Loop tiling'' are running time with loop tiling.
Bars with ``Early termination'' represents timing with both loop tiling and early termination, and those labeled with ``Vectorization'' are timing with all the three optimization methods.
Since there are many synchronizations in Phase 1, it gains a good deal of performance improvement from loop tiling.
Early termination applies only to Phase 2.
Because there are many small posterior values, the early termination is very effective in reducing the time of Phase 2.
%saves significant amount of computation.
Vectorization gave a further improvement for Phase 1, which has a good amount of dense computation.

\begin{algorithm}[htb]
\caption {Basic loop structure of Phase 2 on GPU} 
\label{alg:basic}
\algsetup{indent=1em}
\begin{algorithmic}[1]
\renewcommand{\algorithmicrequire}{\textbf{Input:}}
\renewcommand{\algorithmicensure}{\textbf{Output:}}
\renewcommand{\algorithmicforall}{\textbf{foreach}}
\FOR{$i=0$ to $num\_descriptors$, $step$ = 2}
\FOR{$j=0$ to $num\_components$, $step$ = 4}
\IF{$posterior_{i,j}>threshold$ }
\STATE Compute $U$ and $V$ values 
\ENDIF
\IF{$posterior_{i,j+1}>threshold$ }
\STATE Compute $U$ and $V$ values
\ENDIF
\\$\cdots$
\ENDFOR
\ENDFOR
\end{algorithmic}
\end{algorithm}

\begin{algorithm}[htb]
\caption {Optimized loop structure of Phase 2 on GPU} 
\label{alg:optimized}
\algsetup{indent=1em}
\begin{algorithmic}[1]
\renewcommand{\algorithmicrequire}{\textbf{Input:}}
\renewcommand{\algorithmicensure}{\textbf{Output:}}
\renewcommand{\algorithmicforall}{\textbf{foreach}}
\FOR{$i=0$ to $num\_descriptors$, $step$ = 2}
\FOR{$j=0$ to $num\_components$, $step$ = 4}
\STATE bool $b1$ = $posterior_{i,j}<1e-6$;
\STATE bool $b2$=$posterior_{i,j+1}<1e-6$;
\STATE $\cdots$
\STATE bool $TotalB$=$b1\&b2\&b3...b8$;
\IF{$TotalB >0$}
\STATE continue;
\ENDIF
\IF{$posterior_{i,j}>threshold$}
\STATE Compute $U$ and $V$ values 
\ENDIF
\IF{$posterior_{i,j+1}>threshold$ }
\STATE Compute $U$ and $V$ values 
\ENDIF
\\$\cdots$
\ENDFOR
\ENDFOR
\end{algorithmic}
\end{algorithm}

\section{Experiments}
We evaluated our GPU-FV system in two scenarios. The first scenario is image classification with Fisher vector.
The second part is event/expression detection in video, using Fisher vector generated for each frame.
The testing platform is a dual CPU computer, which has two Intel Xeon E5-2630 v3 CPUs at clock rate of 2.40GHz, with 8 cores on each.
It has an NVIDIA Tesla K40 card, with CUDA 7.0 installed.
We used the same encoding scheme as VLFeat, with 256 GMM components, and 82 dimension dense SIFT descriptors at a stride of 4 on each dimension of the original image.
The 9 scales of an image are selected from 2 to 1/8, with a decreasing factor  of $\sqrt{2}$.
\subsection{Comparing  GPU-FV with Fisher vectors on CPUs}%Image classification}
\begin{figure}[ht]
\includegraphics[width=7cm]{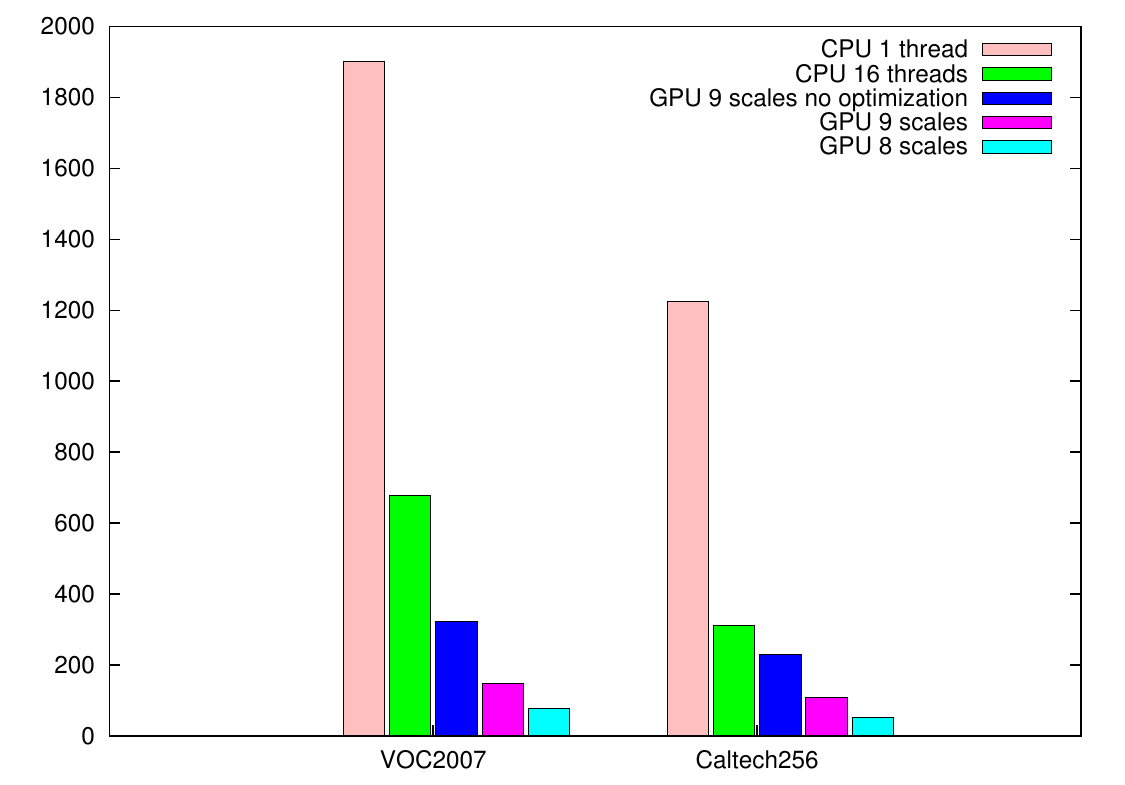}
    %testsfv.pdf}
\caption{Average encoding time for each image}
\label{fig:time}
\end{figure}

% on PASCAL and Caltech256 datasets}
%speed vs accuracy
We tested our GPU code on the PASCAL VOC2007 and Caltech256 image sets. We compare GPU-FV with the MATLAB CPU code in VLFeat.
The same encoding and classification algorithms are used in all the versions.
The average encoding time for each image is plotted in Figure~\ref{fig:time}, and the accuracy values in mAP are listed in Table~\ref{table:accuracy}.
For the CPU code, since VLFeat provided OPENMP implementation, we tested it on 1 thread and 16 threads, shown as the first two bars in each cluster of Figure~\ref{fig:time}.
To show the effectiveness of the optimization to the GPU code, we included tests on GPU with and without the optimizations mentioned in Section~\ref{sec:opt}, plotted as the two bars labeled ``GPU 9 scales no optimization'' and ``GPU 9 scales''.
By ``9 scales'', we mean that dense SIFT is extracted from 9 scales of an image, which is the original setting in VLFeat.
For the GPU code, we also tested the version with 8 scales in dense SIFT by removing the largest scale (resized to 2 times of the image), shown as the last bar in each cluster of the figure.
Using 8 scales and with all the optimization strategies, the average encoding time for each image is 77.58ms for the VOC2007 set, and 53.99ms for the CalTech256 set.
We did this test to show that this approximation (removing one scale) could improve performance dramatically, while maintaining about the same accuracy.

%Test on VOC (5011 images for training, 4952 images for testing)
The tests on VOC2007 used 5011 images for training the svm classifier, and used another 4952 images as testing set.
The average encoding time for each picture is shown in the left cluster of bars in Figure~\ref{fig:time}. 
Our optimized GPU code got a speedup of 12.68 over the 1 thread CPU code, and 4.53 over the 16 thread CPU code.
The optimization of GPU code improves performance by 53.5\% over the unoptimized version.
The tests on 8 scales further reduced the running time by 48\%, but the accuracy is not affected by removing one scale.
The accuracy of each class is shown in Table~\ref{tab:class}.
%Figure~\ref{fig:class}.

%\begin{figure}[ht]
%\centering
%\includegraphics[width=10cm]{class.png}
%\caption{AP of each class in VOC2007}
%\label{fig:class}
%\end{figure}

%\begin{comment}
\begin{table}[htb]
  \caption{The accuracy of each class in VOC2007}
  \begin{center}
    \footnotesize
    \begin{tabular}{lcc}
      {\bf class} & {\bf CPU} & {\bf GPU} \\
      {}& {\bf 9scales } & {\bf 8scales}\\
      {}& {\bf (\%) } & {\bf (\%)}\\
aeroplane    &82.90 &  81.75  \\
bicycle    &65.44     &  65.52 \\
bird & 57.40 & 52.93\\
boat & 72.94 & 72.60 \\
bottle & 27.12 & 28.03 \\
bus & 65.32 & 67.01 \\
car & 81.51 & 81.45 \\
cat & 57.56 & 55.97 \\
chair & 50.92 & 45.89 \\
cow & 43.60 & 43.45 \\
diningtable & 58.00 & 55.51 \\
dog & 41.84 & 41.05 \\
horse & 81.84 & 81.07 \\
motorbike & 66.64 & 68.67 \\
person & 84.97 & 84.51 \\
pottedplant & 30.19 & 27.88 \\
sheep & 48.30 & 46.33 \\
sofa & 56.70 & 53.56 \\
train & 82.07 & 81.63 \\
tvmonitor & 52.98 & 51.16 \\
\bottomrule
%\caption{AP of each class in VOC2007}
\end{tabular}
    \label{tab:class}
    \end{center}
\end{table}

%\end{comment}

\begin{table}[htb]
\caption{The accuracy of different versions}
\begin{center}
\footnotesize
\begin{tabular}{lccc}
   {\bf Data set} & {\bf CPU} &  {\bf GPU} & {\bf GPU }\\
                 & {} &{\bf 9scales }   & {\bf 8scales}\\
%\midrule
VOC2007 (mAP) & 59.93\% & 59.03\%  & 59.3\% \\

Caltech256 (accuracy) & 42.08\% & 42.88\% & 39.48\%\\
\bottomrule
\end{tabular}
\label{table:accuracy}
\end{center}
\end{table}

%1 CPU: 1.9s for each picture.
%18953.93s for encoding, 
%16 threads: 678.68ms for each picture.
%710.51s for classification. 
%mAP: 59.93 \%

%8 CPU workers: about 145 seconds per chunk, about 2900s for total encoding:
%12 CPU workers: about 114 seconds per chunk, about 2280s for total encoding:

%1 GPU:
%8 scales: 126.832s for dsift, 679.603s for fisher vector. 806.435s for total encoding. 
%mAP: 59.30 \%
%9 scales: 200.143s for dsift, 1293.88s for fisher vector. 1494.023s for total encoding, 4.107s for classification.
%706.059 for classification.
%mAP: 59.03 \%

%1GPU+8 threads
%9 scales: 1506.46s for total encoding.

The test on Caltech256 uses 7680 images for training, and 6400 images for testing, with more diversity in image size and shape. 
The performance is shown in the 5 bars on the right in Figure~\ref{fig:time}.
The GPU code got a speedup of 21.8 over the 1 thread version, and 5.53 over the 16 thread version.
The optimization reduced average running time on 9 scales GPU code by 52.5\%.
The accuracy on 8 scales is still acceptable, while the time is reduced by 50\%.

%1 CPU: 1.23s for each picture.
%9 scales: 
%17306.65s for encoding (including writing files), 1615.82 for classification.
%mean accuracy 42.08\%.

%16 threads: 310.79ms for each picture.
%1 GPU: 
%9 scales:302.495s for dsift, 1228.87s for fisher vector. 1531.365s for total encoding, 1582.63s for classification.
%mean accuracy: 42.88 \%
%8 scales:  217.086 for dsift, 573.957s for fisher vector. 791.043 for total encoding, 1619.32s for classification.
%mean accuracy: 39.48 \%

\subsection{Applications for Real Time Video Monitoring}

In our tests with Caltech256 set, we found that the encoding time on GPU is a few dozens of milli-seconds for small pictures.
Therefore, we try to apply GPU-FV to real-time video processing scenarios.
By encoding each frame in a video,  various classification and detection jobs could be conducted.
We show two experiments on video processing in this paper.
In detecting abnormal events, we used the GMM components trained in the Caltech256 experiment to generate Fisher vectors for the frames.
For baby laughing detection, we used GMM components trained from random samples in pubfig, which is a data set with pictures of human faces.
%abandon bags

\begin{figure}[ht]
\begin{minipage}{0.45\linewidth}
\includegraphics[width=3.8cm]{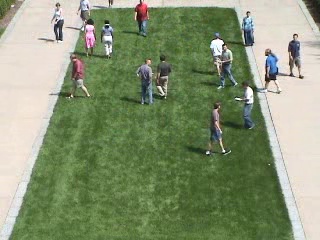}
\end{minipage}
\begin{minipage}{0.45\linewidth}
\includegraphics[width=3.8cm]{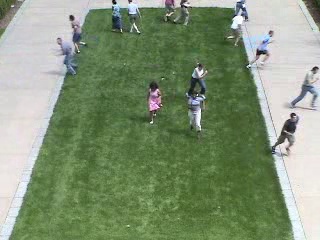}
\end{minipage}
\caption{Example frames for UMN abnormal event dataset.}
\label{fig:abnormal_event}
\end{figure}

%Abnormal event:
In the first experiment, we tried to detect abnormal events from the UMN crowd activity video\footnote{http://mha.cs.umn.edu/}.
%http://mha.cs.umn.edu/Movies/Crowd-Activity-All.avi 
The video consists of 3 scenes. 
In each scene, people are walking casually or standing.
At a certain time, all the people are running away in various directions.
The three scenes are in different places, with different light conditions.
Each frame has a label to denote whether it is in the abnormal event.
The frames with people running are labeled as ``abnormal events''.
Figure~\ref{fig:abnormal_event} shows a frame without any abnormal event on the left, and a scene with abnormal event on the right.
The resolution of the video is 320*240 pixels.

In each scene, we used the first 700 frames to train a classification model with liblinear~\cite{liblinear}.
Then we use the model to predict abnormal events in the scene.
%The frames are of size 320*240, and the average encoding time is about 37ms for each frame. 

In the experiments, we used 8 scales for dense SIFT descriptors, and the total encoding time for each
frame is 34ms on average. It means that we can process about 29 frames in a second, implying that we can apply the classification to real-time usage.
%It means that we can process about 29 frames in a second.
We compare the performance of our GPU-FV with some well-known event detection methods in Table~\ref{table-surveillance}, with the last column listing the time for generating feature for each frame (or for each segment for C3D).
C3D generates feature for every segment of 16 frames, so we evaluated its AUC with prediction for both segment level and frame level.
%We present the result of C3D using $pool5$ net, which got the best prediction score among all of the networks it provided. 
It can be seen that our GPU-FV is much faster than the traditional event detection methods.
%, which are shown in Table~\ref{table-surveillance}. 
%The processing will be even faster with lower resolution.
%With this speed, we can apply the classification to real-time usage.

\begin{table}[htb]
	\caption{The performance on UMN dataset (three scenes together). }
% The last column is the time for encoding a segment (C3D) or a frame (all the other methods). C3D does the encoding with a segment of 16 frames. We evaluated C3D at segment level (the 6th row), and at frame level (the 7th row).}
	\begin{center}
		\footnotesize
		\begin{tabular}{lcc}
			\hline
			{\bf Method} & {\bf AUC} &  {\bf Encoding Time} \\
			\hline

			%Our method & 0.982 & about 0.034 s  \\
			Our GPU-FV & 0.984 &  0.034s   \\
			Deep CNN feature \cite{jia2014caffe} & 0.930 & 0.020s  \\
			sparse reconstruction cost \cite{bib-src} & 0.978 & 0.8s \\
			local statistical aggregates \cite{saligrama2012video}  &0.985 & 1.1s \\
			social force \cite{mehran2009abnormal} &0.960 & 5s \\
                        C3D \cite{DuTran15ICCV} &0.945(segment)  & 0.053s \\
                        C3D \cite{DuTran15ICCV} &0.946(frame)  & 0.053s\\
			\hline
		\end{tabular}
		\label{table-surveillance}
	\end{center}
\end{table}

\begin{figure}[ht]
	%  \begin{minipage}
	\begin{minipage}{0.48\linewidth}
		\includegraphics[width=4.4cm]{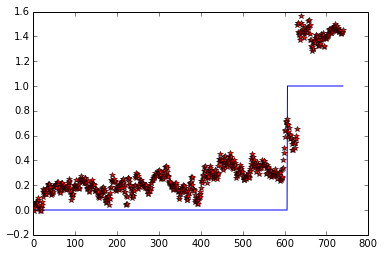}
	\end{minipage}
	\begin{minipage}{0.48\linewidth}
		\includegraphics[width=4.4cm]{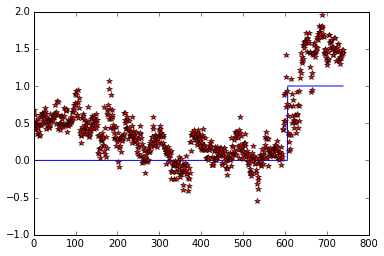}
	\end{minipage}
	
	\begin{minipage}{0.48\linewidth}
		\includegraphics[width=4.4cm]{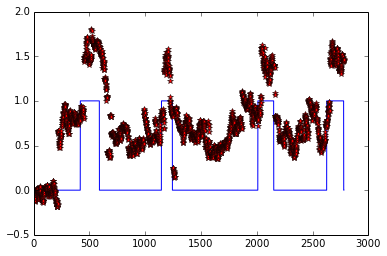}
	\end{minipage}
	\begin{minipage}{0.48\linewidth}
		\includegraphics[width=4.4cm]{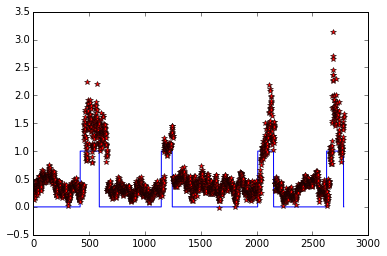}
	\end{minipage}

	\begin{minipage}{0.48\linewidth}
		\includegraphics[width=4.4cm]{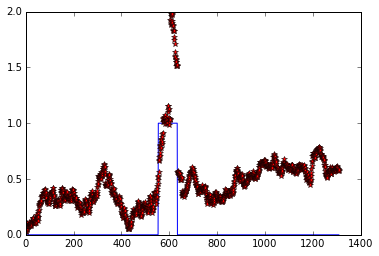}
	\end{minipage}
	\begin{minipage}{0.48\linewidth}
		\includegraphics[width=4.4cm]{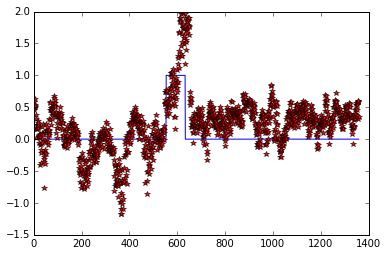}
	\end{minipage}
	%\end{minipage}
	\caption{Comparing the prediction values using GPU-FV (left) and deep CNN feature (trained from ImageNet using AlexNet). }
	\label{fig:abnormal-compare}
\end{figure}

Deep CNN features with Caffe is the only one that outperforms GPU-FV in encoding time, but at a lower AUC value.
Figure~\ref{fig:abnormal-compare} compares the prediction values using GPU-FV and deep CNN features \cite{DBLP:journals/corr/RazavianASC14} on the three scenes.
We presented the prediction values from liblinear with the red star lines. The blue lines depict the ground truth, in which a value ``1'' represents an abnormal event.
In the 3 graphs on the left, which are results using GPU-FV, it can be seen that the bursts with high value in the red start lines are consistent with the periods in which the blue lines are with value ``1''. 
And it is clear that GPU-FV outperforms deep CNN features by providing clearer distinction between normal scenes and abnormal scenes.
The average AUC (area under ROC curve) for GPU-FV is 0.984, while the average AUC for deep features is 0.930. The reason why deep feature does not work well for this problem is because the number of training examples in the classification of frames in a video is too small. In addition, the scenario of abnormal event detection is very different from ImageNet objects, where the deep CNN features are trained from. 

%Figure~\ref{fig:abnormal1} shows the prediction values of the 3 scenes, with each row represents a scene. The first picture in each row depicts the prediction value of the first 700 frames, and the second picture depicts the other frames in the scene.
%There are two abnormal events in the first scene, which can be seen obviously from the prediction values in the figure.
%The AUC (area under ROC curve ) is 0.998 on the 740 frames which serve as the testing set.
%The abnormal events can also be observed from the next four pictures, which plot the prediction values of Scene 2 and Scene 3.
%It can be seen that all the abnormal events can be detected in Scene 2 and Scene 3.
%The average AUC is 0.984.
%We tested this application with deep feature~\cite{jia2014caffe}, and present the results in Figure~\ref{fig:abnormal2}, which has AUC=0.93.
%It can be seen that our results with Fisher vector is better than deep feature.

\begin{figure}[ht]
\begin{minipage}{0.45\linewidth}
\includegraphics[width=3.8cm]{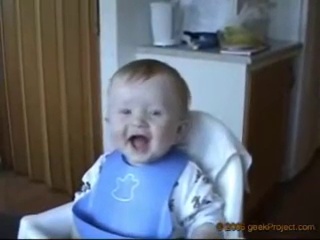}
\end{minipage}
\begin{minipage}{0.45\linewidth}
\includegraphics[width=3.8cm]{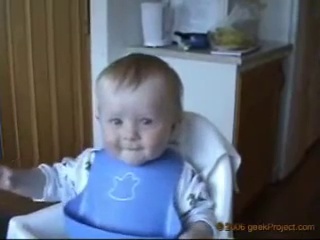}
\end{minipage}
\caption{Example frames of baby expression video.}
\label{fig:babysmile}
\end{figure}

Another experiment we did is detecting baby's laugh from a video. The video is also of resolution 320*240.
We labeled the frames when a baby is laughing as 1, and the other frames as 0.
The left picture in Figure~\ref{fig:babysmile} shows a frame when the baby is laughing, the one on the right is a frame when the baby is not laughing.
Similar to the first experiment, 
we used the first 700 frames as training set, and used the following frames as testing set.
%The prediction values of the frames in the training set and those in the testing set are shown in the two pictures in Figure~\ref{fig:baby}.
%It can be seen that the laughing periods could be distinguished by examining the prediction values.
%We also tested it with deep feature, and the results are shown in Figure~\ref{fig:babysmile-deep}.
%It is obvious that our results are better than those with deep feature.
%The prediction values of the frames in the training set and those in the testing set are shown in the two pictures in Figure~\ref{fig:baby}.
%The AUC in the test set is 0.935.
%It can be seen that the laughing periods could be distinguished by examining the prediction values.
We also tested this video using CNN from Caffe \cite{jia2014caffe} and C3D \cite{DuTran15ICCV}, with performance comparison listed in Table~\ref{table-baby}.
It shows that though encoding on Caffe is faster, our method obtained a much higher accuracy than CNN features (0.935 compared to 0.655).
And because we extracted features for each frame, we got a much higher AUC than C3D when evaluated at frame level.

%\begin{figure}[ht]
%\begin{minipage}{0.48\linewidth}
%\includegraphics[width=4.4cm]{predict_scene1_1.pdf}
%\includegraphics[width=4.4cm]{baby1.png}
%\end{minipage}
%\begin{minipage}{0.48\linewidth}
%\includegraphics[width=4.4cm]{predict_scene1_2.pdf}
%\includegraphics[width=4.4cm]{baby2.png}
%\end{minipage}
%\caption{Prediction values of each frame for testing baby laughing with Fisher vector}
%\label{fig:baby}
%\end{figure}
\begin{table}[htb]
	\caption{The performance on detecting baby's laugh.}
	\begin{center}
		\footnotesize
		\begin{tabular}{lcc}
			\hline
			{\bf Method} & {\bf AUC} &  {\bf Encoding time} \\
			\hline

			%Our method & 0.982 & about 0.034 s  \\
			Our GPU-FV & 0.935 & 0.034 s  \\
			Deep CNN feature \cite{jia2014caffe} &0.655 & 0.020 s\\
                        C3D \cite{DuTran15ICCV} &0.914(segment) & 0.083 s\\
                        C3D \cite{DuTran15ICCV} &0.593(frame) & 0.083 s\\
			\hline
		\end{tabular}
		\label{table-baby}
	\end{center}
\end{table}

%\begin{figure}[ht]
%\begin{minipage}{0.48\linewidth}
%\includegraphics[width=4.4cm]{predict_scene1_1.pdf}
%\includegraphics[width=4.4cm]{pic/smile_deep_tr.png}
%\end{minipage}
%\begin{minipage}{0.48\linewidth}
%\includegraphics[width=4.4cm]{predict_scene1_2.pdf}
%\includegraphics[width=4.4cm]{pic/smile_deep_te.png}
%\end{minipage}
%\caption{Prediction values of each frame for testing baby laughing with deep feature}
%\label{fig:babysmile-deep}
%\end{figure}

\section{Conclusion}
In this paper, we introduced an optimized implementation of Fisher vector on GPU (GPU-FV), and showed its application to image classification and events detection in videos.
Our method demonstrated a promising approach to using Fisher vector for real-time video processing. In future, we plan to expand the algorithm to more applications in video processing, to provide support for real-time situations.

%ACKNOWLEDGMENTS are optional
\section{Acknowledgments}
This work is supported by the National High-tech R\&D Pro-
gram of China (No. 2012AA 010902), and the National Natural
Science Foundation of China (No. 61303059).
\bibliographystyle{abbrv}
\bibliography{fv}  % sigproc.bib is the name of the Bibliography in this case
% ACM needs 'a single self-contained file'!
%\balancecolumns
% This next section command marks the start of
% Appendix B, and does not continue the present hierarchy
%\balancecolumns % GM June 2007
% That's all folks!
\end{document}